\pdfoutput=1
\documentclass{article}

\usepackage[numbers]{natbib}
\usepackage{arxiv}
\usepackage{multirow} 
\usepackage[utf8]{inputenc} 
\usepackage[T1]{fontenc}    
\usepackage{hyperref}       
\usepackage{url}            
\usepackage{booktabs}       
\usepackage{amsfonts}       
\usepackage{nicefrac}       
\usepackage{microtype}      
\usepackage{cleveref}       
\usepackage{lipsum}         
\usepackage{graphicx}
\usepackage{natbib}
\usepackage{doi}
\usepackage{algorithm}
\usepackage{algorithmic}

\title{From Images to Decisions: Assistive Computer Vision for Non-Metallic Content
Estimation in Scrap Metal}

\newif\ifuniqueAffiliation
\uniqueAffiliationtrue

\ifuniqueAffiliation 
\author{ \href{https://orcid.org/0009-0007-1647-8868}{\includegraphics[scale=0.06]{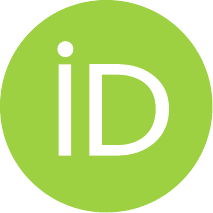}\hspace{1mm}Daniil Storonkin} \\
	Department of Computer Science\\
	ITMO University, Saint Petersburg, Russia\\
	ISP RAS, Moscow, Russia \\
	\texttt{storonkin.2k17@gmail.com} \\
	\And
	\href{https://orcid.org/0000-0000-0000-0000}{\includegraphics[scale=0.06]{orcid.pdf}\hspace{1mm}Ilia Dziub} \\
	Independent researcher\\
	\texttt{dzyubilya@gmail.com} \\
	\AND
    \href{https://orcid.org/0000-0002-0679-6981}{\includegraphics[scale=0.06]{orcid.pdf}\hspace{1mm}Maksim Golyadkin} \\
	Department of Computer Science\\
	AIRI, Moscow, Russia \\
	\texttt{mxmgolyadkin@gmail.com} \\
	\And
    \href{https://orcid.org/0000-0002-3308-8825}{\includegraphics[scale=0.06]{orcid.pdf}\hspace{1mm}Ilya Makarov} \\
	Department of Computer Science\\
    ITMO University, Saint Petersburg, Russia\\
Research Center of the Artificial Intelligence Institute, \\ Innopolis University, Innopolis, Russia\\
	AIRI, Moscow, Russia\\
    ISP RAS, Moscow, Russia \\
	\texttt{iamakarov@hse.ru} \\
}
\else
\fi

\renewcommand{\headeright}{}
\renewcommand{\undertitle}{Technical Report}

\hypersetup{
pdftitle={A template for the arxiv style},
pdfsubject={q-bio.NC, q-bio.QM},
pdfauthor={David S.~Hippocampus, Elias D.~Striatum},
pdfkeywords={First keyword, Second keyword, More},
}

\begin{document}
\maketitle

\begin{abstract}
Scrap quality directly affects energy use, emissions, and safety in steelmaking. Today, the share of non-metallic inclusions (contamination) is judged visually by inspectors - an approach that is subjective and hazardous due to dust and moving machinery. We present an assistive computer vision pipeline that estimates contamination (per percent) from images captured during railcar unloading and also classifies scrap type. The method formulates contamination assessment as a regression task at the railcar level and leverages sequential data through multi-instance learning (MIL) and multi-task learning (MTL). Best results include MAE 0.27 and R2 0.83 by MIL; and an MTL setup reaches MAE 0.36 with F1 0.79 for scrap class. Also we present the system in near real time within the acceptance workflow: magnet/railcar detection segments temporal layers, a versioned inference service produces railcar-level estimates with confidence scores, and results are reviewed by operators with structured overrides; corrections and uncertain cases feed an active-learning loop for continual improvement. The pipeline reduces subjective variability, improves human safety, and enables integration into acceptance and melt-planning workflows.
\end{abstract}


\section{Introduction}

Scrap metal use in steel production is increasing annually, improving environmental sustainability by reducing reliance on virgin resources and reducing production costs. However, the quality of scrap can vary due to contaminants, coatings, and impurities, which can degrade the properties of recycled metal, cause defects, and increase processing expenses. Accurate and efficient define contamination is essential to ensure high-quality materials and maintain the integrity of the recycling process ~\cite{nmis2021,cis2018}.

\textbf{Impact of Non-Metallic Inclusions on Processing Efficiency and Energy Consumption.} Non-metallic inclusions in scrap metal, adversely affect the Electric Arc Furnace (EAF) recycling process's efficiency and increase energy consumption ~\cite{klimek2024circulartransformationeuropeansteel}. These impurities degrade molten metal quality, reduce steel yields, and necessitate additional slag formation, disrupting continuous melting and lowering productivity~\cite{ImpactNMI}.

\begin{figure*}[t]
\centering
\includegraphics[width=1.005\textwidth]{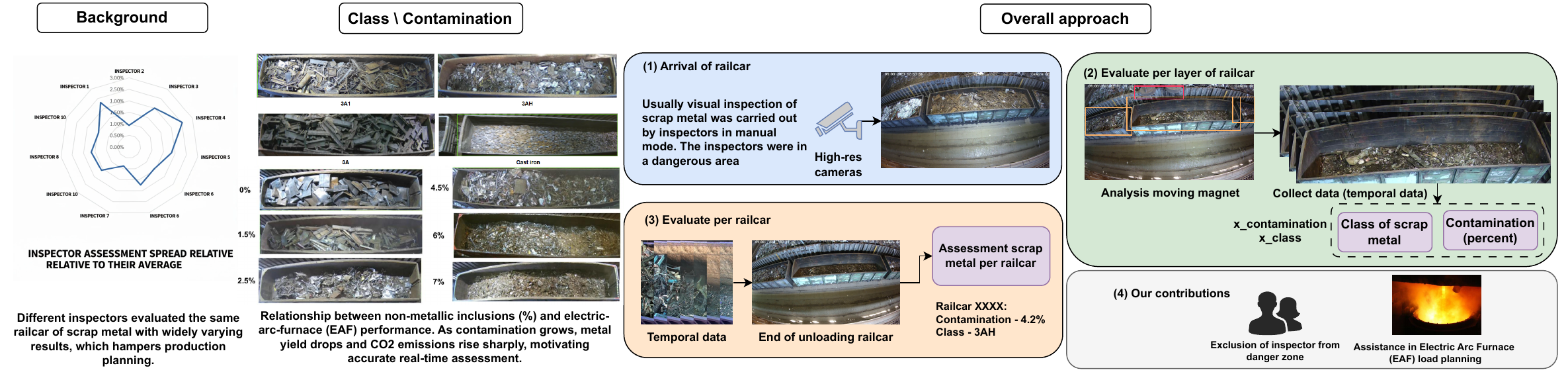} %
\caption{\textbf{Background}: The determination of non-metallic inclusions in scrap metal depends on the inspector’s subjective judgment, which reduces the quality of production planning. The chart shows that assessments of the same railcar by different inspectors vary significantly. \textbf{Class \& Contamination: }During a scrap metal assessment, the inspector handles several tasks at once: determining the percentage of contamination in the scrap and its grade. \textbf{Overall approach}: railcar arrives with scrap metal at the unloading point under the camera. The detector analyzes the location of the magnet relative to the railcar. There used to be an inspector in the unloading area who was exposed to airborne dust and other hazards.}
\label{flow}
\end{figure*}

\textbf{Safety and Subjectivity Issues in Manual Contamination Assessment.} Workers involved in unloading scrap metal face hazards from strong magnetic fields, dust, and particulate matter, increasing the risk of accidents and respiratory issues. Additionally, relying on human inspectors introduces subjectivity, leading to inconsistent contamination reports and unreliable quality control (Figure ~\ref{flow} - Background). Variability in inspectors' criteria can cause economic inefficiencies, such as increased processing or rejection of acceptable scrap.

To date traditional methods of visual control rely on manual sorting and inspection, which are time-consuming, labor-intensive, and prone to errors, making it difficult to handle increasing scrap volumes. To solve these problems, we propose a \textbf{first deep learning pipeline} to automate and improve the definition of contamination process (Figure~\ref{flow} - Overall approach).

Thus, our main contributions are as follows:

\begin{itemize}
    \item We propose methods based on MIL and MTL and show that they achieve quality metrics comparable to those of an inspector.
    \item We conducted experiments on our dataset in both single-task mode (separate define of contamination and class scrap metal) and multi-task mode.  This task helps processors optimize quality control, reduce operating costs, and minimize waste.
    \item  We demonstrate that transformer-based models significantly outperform traditional CNN architectures in contamination prediction.

\end{itemize}

\begin{figure*}[t]
\centering
\includegraphics[width=1.02\textwidth]{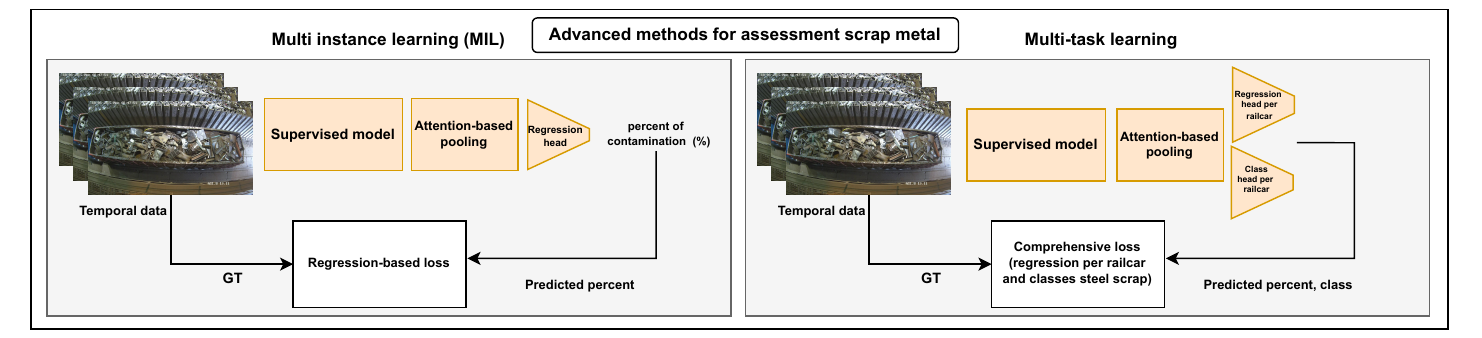} 
\caption{Differences between variants of deep learning architectures. Multi-instance learning approach takes a temporal data (a bag) and,
based on multiple instances, assigns a score to the railcar. With multi-task learning, we solve several tasks for scrap metal
assessment in a single inference by using multiple heads.}
\label{methods}
\end{figure*}

\section{Related work}

\textbf{Scrap Metal Processing.} The authors~\cite{smirnov2020,smirnov2021} adapt the methods of machine learning and deep learning for metal scrap classification, also adding the automatic cropping of metal scrap  images from photographs of railway carriages, significantly simplifying the data preparation for classification. The article ~\cite{linh2019} proposes a system for the automatic classification of metal waste, aimed at reducing the time and effort required for manual sorting using simple neural networks. 

~\cite{auer2019artificialintelligencebasedprocessmetal} propose an AI-based approach to identify tool and high-speed steel alloys from electric-arc optical emission spectra, benchmarking several classifiers and finding a linear SVM to perform best.

A cascade machine learning technique ~\cite{balakrishnan2018} is used for garbage detection and classification, the essence of which is to train a universal detector and classification based on SVM. 

We are the first to address the scrap-metal assessment problem while simultaneously performing comprehensive scrap analysis: evaluating non-metallic inclusions in scrap from multiple railcar images and determining the scrap grade.

\section{Methods}

\subsection{Multi Instance Learning for percentage of scrap metal contamination at railcar level}

\begin{table}[ht]
    \centering
    \begin{tabular}{llcc}
        \hline
         & Model & $\mathrm{MAE}_{railcar}$ & $R^2_{railcar}$ \\
        \hline
        \multirow{2}{*}{\textbf{Models}}
            & EfficientNet 7 B  & 1.43 & -0.23 \\
            & ResNet 50         & 0.73 & 0.06  \\
            & ResNeXt 101       & 0.56 & 0.45  \\
            & ViT 16 b          & 0.56 & 0.40  \\
            & ViT 32 b          & 0.61 & 0.31  \\
            & \textbf{Swin 2}   & \textbf{0.27} & \textbf{0.83} \\
        \hline
        \multirow{2}{*}{\textbf{Inspector}}
            & Inspector Avg     & $0.19 \pm 0.012$  & 0.93 \\
            & Inspector Best    & \textbf{0.18} & \textbf{0.94} \\
        \hline
    \end{tabular}
    \caption{Effectiveness of multi-instance learning (MIL) and inspectors
    for railcar-level contamination. Swin-MIL with a bag of 5 layers attains MAE = 0.27\% and $R^2 = 0.83$. Metrics of best model approach the assessment indicators of inspector.}
    \label{tab:mil}
\end{table}
\begin{table}[ht]
    \centering
    \begin{tabular}{lccccc}
        \hline
        Model &  $\mathrm{R}_{railcar}$ &
    $\mathrm{A}_{railcar}$ &
    $\mathrm{P}_{railcar}$ &
    $\mathrm{F1}_{railcar}$ \\
    \hline
        EfficientN   & 0.60 & 0.61 & 0.63  & 0.61 \\
        ResNet     & 0.53 & 0.56 & 0.57  & 0.55 \\
        ResNeXt    & 0.59 & 0.61 & 0.65  & 0.63 \\
        ViT   & 0.61 & 0.63 & 0.64  & 0.64 \\
        \textbf{Swin 2} & \textbf{0.68} & \textbf{0.73 }& \textbf{0.70}  & \textbf{0.73} \\
        \hline
    \end{tabular}
    \caption{Scrap-grade classification accuracy. Recall (R), Accuracy (A), Precision (P). Transformer architectures (ViT and Swin) exceed 70\% accuracy, surpassing the best CNN by 12 pp, which confirms the importance of long-range context for distinguishing similar scrap types}
    \label{tab:classes_comprasion}
\end{table}
The target percent contamination exists only at the railcar level, while observations arrive as multiple layers during unloading. During unloading, inspectors observe multiple magnet grabs but provide a single, holistic contamination assessment for the entire railcar rather than rating each layer separately. As a result, ground truth exists only at the railcar level, and the model must aggregate per-layer evidence to match the inspector’s decision.

Content is heterogeneous across layers, their number varies, and a few highly contaminated layers can be diluted by naive averaging. Per-grab training also suffers from instance–label mismatch.
Multi-Instance Learning ~\cite{DBLP:journals/corr/abs-1107-2021} is an approach in which a model is trained on sets of instances ("bags" or temporal data) instead of individual examples (See details in Algorithm 1). In the context of assessing the contamination of scrap metal, each railcar is considered a “bag” containing several layers (magnet grabs), each with its own contamination signal (Figure ~\ref{methods} - MIL).
Our results show that concatenating latent features from different layers of the railcar improves the metrics (Table ~\ref{tab:mil}). The SWIN 2 model achieves higher quality than the other models.

\subsection{Multi Task Learning (percent of contamination, scrap metal classificaton) for scrap metal}

\begin{algorithm}[tb]
\caption{MIL Training}
\label{alg:mil}
\textbf{Input}: Dataset $\mathcal{D}=\{(B_i, y_i)\}_{i=1}^N$\\
\textbf{Parameter}: epochs $T$, samples per bag $s$, loss $\mathcal{L}$, optimizer Opt\\
\textbf{Output}: Trained parameters $\Theta$
\begin{algorithmic}[1]
\STATE Initialize parameters $\Theta$ of Encoder, Attention, and Head.
\FOR{$t=1$ to $T$}
  \FOR{each minibatch $\{(B_k, y_k)\}_{k=1}^m$}
    \STATE For each $B_k$, select $s$ instances $X_k=\{x_{k,j}\}_{j=1}^s$.
    \STATE Compute features $f_{k,j} = \mathrm{Encoder}(x_{k,j})$ for all $j$.
    \STATE Compute scores $a_{k,j} = \mathrm{Attention}(f_{k,j})$.
    \STATE Compute weights $\alpha_{k,j} = \mathrm{softmax}(\{a_{k,j}\}_{j=1}^s)$.
    \STATE Aggregate $z_k = \sum_{j=1}^s \alpha_{k,j} f_{k,j}$.
    \STATE Predict $\hat{y}_k = \mathrm{Head}(z_k)$.
    \STATE Compute loss $\ell = \mathcal{L}(\{\hat{y}_k\}, \{y_k\})$.
    \STATE Zero gradients; backpropagate $\nabla_\Theta \ell$; update $\Theta$ with Opt.
  \ENDFOR
\ENDFOR
\STATE \textbf{return} $\Theta$
\end{algorithmic}
\end{algorithm}

\begin{algorithm}[tb]
\caption{MTL Training}
\label{alg:mil_mtl}
\textbf{Input}: Dataset $\mathcal{D}=\{(B_i, y_i^{reg}, y_i^{cls})\}_{i=1}^N$\\
\textbf{Parameter}: epochs $T$, samples per bag $s$, losses $\mathcal{L}_{reg}$, $\mathcal{L}_{cls}$, weight $\lambda_{cls}$, optimizer Opt\\
\textbf{Output}: Trained parameters $\Theta$
\begin{algorithmic}[1]
\STATE Initialize parameters $\Theta$ of Encoder, Attention, Regressor, and Classifier.
\FOR{$t=1$ to $T$}
  \FOR{each minibatch $\{(B_k, y_k^{reg}, y_k^{cls})\}_{k=1}^m$}
    \STATE For each $B_k$, select $s$ instances $X_k=\{x_{k,j}\}_{j=1}^s$ (uniform).
    \STATE Compute features $f_{k,j} = \mathrm{Encoder}(x_{k,j})$ for all $j$.
    \STATE Compute scores $a_{k,j} = \mathrm{Attention}(f_{k,j})$.
    \STATE Compute weights $\alpha_{k,j} = \mathrm{softmax}(\{a_{k,j}\}_{j=1}^s)$.
    \STATE Aggregate $z_k = \sum_{j=1}^s \alpha_{k,j} f_{k,j}$.
    \STATE Predict $\hat{y}_k^{reg} = \mathrm{Regressor}(z_k)$ and $\hat{y}_k^{cls} = \mathrm{Classifier}(z_k)$.
    \STATE Compute loss $\ell = \mathcal{L}_{reg}(\hat{y}_k^{reg}, y_k^{reg}) + \lambda_{cls}\,\mathcal{L}_{cls}(\hat{y}_k^{cls}, y_k^{cls})$.
    \STATE Zero gradients; backpropagate $\nabla_\Theta \ell$; update $\Theta$ with Opt.
  \ENDFOR
\ENDFOR
\STATE \textbf{return} $\Theta$
\end{algorithmic}
\end{algorithm}

The inspector is handling several tasks at once (estimate contamination level and scrap metal class), so we evaluate the ability of the models to solve multiple tasks simultaneously. Operations require both a numeric contamination estimate and a scrap-grade label. The tasks are correlated (grade frames typical appearance and contamination ranges; contamination cues help grade boundaries). Training separate models wastes data/compute and can yield inconsistent outputs.
Multi-Task Learning ~\cite{10.1023/A:1007379606734} is a machine learning method in which a model simultaneously solves several tasks using common features to increase efficiency and accuracy. In the context of scrap metal analysis, MTL allows you to combine two interrelated tasks: determination of the percentage of scrap metal contamination and classification of scrap metal types (Figure ~\ref{methods} - Multi-task learning). We optimize a joint objective $\mathcal{L}=\mathcal{L}_{reg} + \lambda  \mathcal{L}_{cls}$ with a shared backbone and lightweight task heads, selecting $\lambda$ on the validation set to balance loss scales and stabilize training (See details in Algorithm 2). In practice, joint learning improves sample efficiency and calibration while reducing inference latency compared to maintaining two separate single-task models.

The performance of the evaluated methods is presented in Table ~\ref{tab:mlt}. On average, the metrics of all model architectures except SWIN are better than those of the MIL approach from section MIL. Models such as ResNet and EffNet perform worse on both contamination assessment and scrap metal classification. Transformer-based architectures are better suited to multitask approaches. Despite these advances, much remains to be done to improve accuracy.
\begin{figure}[!tbp]
\centering
\includegraphics[width=0.9\textwidth]{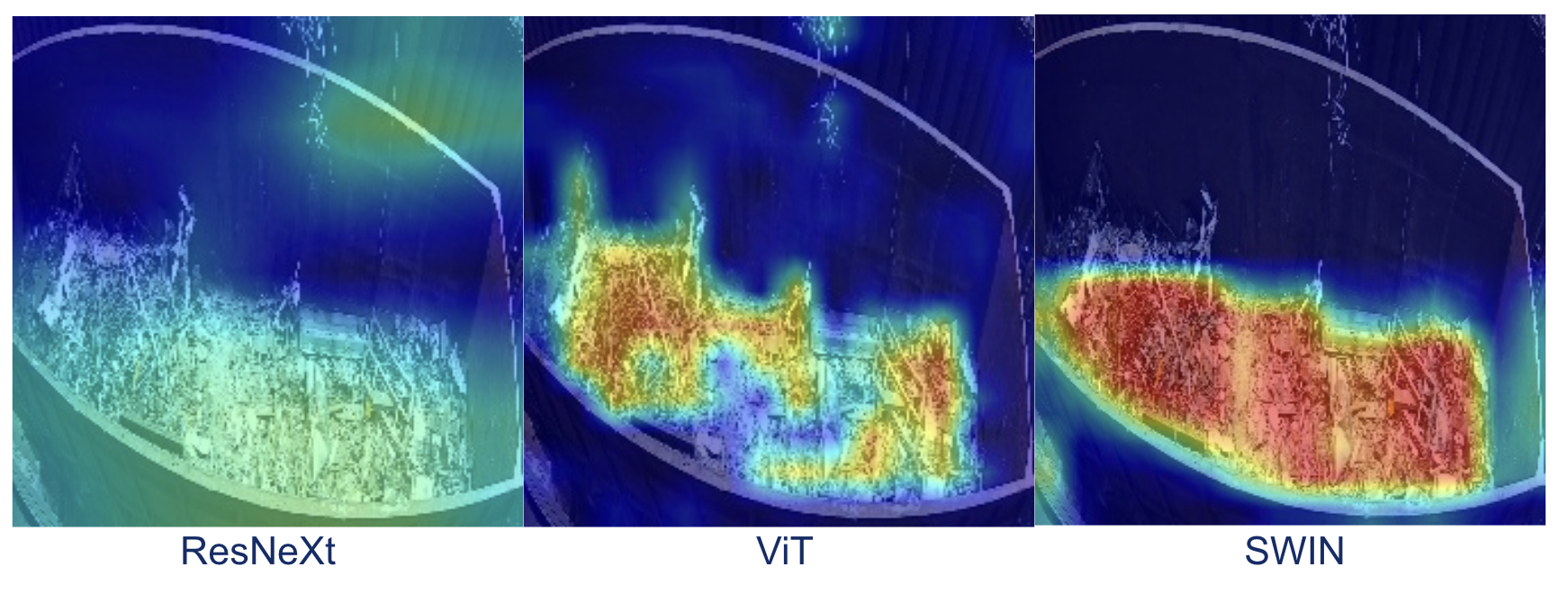}
\caption{Grad-CAM visualisations for CNN vs transformer backbones. Transformers (right) focus tightly on scrap pieces, ignoring dust clouds and background, which explains their superior regression accuracy.}
\label{fig:gradcam}
\end{figure}
\begin{table*}
\centering
  \begin{tabular}{lcccccccc}
    \hline
    Model  & $\mathrm{MAE}_{railcar}$ & $R2_{railcar}$\  &
    $\mathrm{Recall}_{railcar}$ &
    $\mathrm{Acc}_{railcar}$ &
    $\mathrm{Precision}_{railcar}$ &
    $\mathrm{F1}_{railcar}$ \\
    \hline
    EfficientNet 7 B & 0.80 & 0.45 & 0.52 & 0.54 & 0.51 &  0.52 \\
    ResNeXt 101 & 0.55 & 0.58 & 0.62 & 0.67 & 0.53 &  0.56 \\
    ResNet 50 & 0.74 & 0.49 & 0.51 &  0.27 & 0.41 & 0.43 &\\
    VIT B 16 & 0.48 & 0.66 & 0.65 & 0.68 & 0.68 & 0.64 \\
    VIT B 32 & 0.49 & 0.65 & 0.66 & 0.70 & 0.67 & 0.63 \\
    \textbf{Swin 2 B}  & \textbf{0.36} & \textbf{0.78} & \textbf{0.79}  &  \textbf{0.80} & \textbf{0.80} & \textbf{0.79} \\
    \hline
\end{tabular}
  \caption{Multi-task learning (MTL): joint contamination regression + grade classification. The Swin-MTL model reaches state-of-the-art results on both tasks simultaneously (MAE = 0.36\%, R2 = 0.78, F1 = 0.79), showing that shared representations boost performance compared to single-task or MIL baselines.}
  \label{tab:mlt}
\end{table*}

\section{Experimental setup}
\begin{table*}
\centering
    \begin{tabular}{lrr}
        \hline
        Split dataset  & Count layers & Count railcars \\
        \hline
        Train     & 44 092          & 1504        \\
        Val  & 8 575          & 305        \\
        Test & 5 907          & 223        \\

        \hline
    \end{tabular}
    \caption{Our dataset: We have collected
data on the handling of over 90,000 tons of metal scrap (over 2000
railcars)}
    \label{tab:info_dataset}
\end{table*}
In our experiments, we compared several deep learning models that differ conceptually (CNN vs Transformer) and in parameter count, to predict scrap metal contamination levels from images.
The models evaluated include CNN architectures such as EfficientNet B ~\cite{effnet2019}, ResNet 50 ~\cite{resnet2015}, ResNeXt 101 ~\cite{resnext2016}, and transformer-based models like Vision Transformer (ViT B) ~\cite{vit2020} and Swin Transformer (Swin 2 B) ~\cite{swin2021}, all evaluated architectures were with a pre-trained imagenet1K ~\cite{imagenet} weighting base.

\textbf{Dataset.} Our dataset contains \textbf{58,574}  annotated samples (from 40 cameras), with a total of \textbf{2000} railcars (Table \ref{tab:info_dataset}).
According to the acceptance data (Figure ~\ref{flow} - Background), the spread of individual inspectors relative to consensus reaches up to ~2-3 percentage points, while the most stable ones have less than 0.5–1.0 percentage points. This confirms the presence of both an individual offset and a different "width" of the scale for inspectors.
To minimize individual subjective biases, each sample in our dataset is annotated by three independent annotators, and we aggregate their scores by taking the mean. This approach ensures a more reliable and balanced dataset by aggregating multiple perspectives. After annotation, we compute the standard deviation (STD) across the three annotators to assess consistency. If STD exceeds 0.4, the sample is flagged for additional verification (For details, see subsection "Double-blind annotation system"). For the classification task, we also used multiple inspectors. This task is easier to label because there are numerous standards for determining scrap gradations (depending on the geometry and size of the waste). Our dataset contains three classes (3A, 3A1, 3AH) of ferrous scrap and one class of cast iron. We split the dataset by railcars to avoid any train/validation/test leakage: all layers from the same railcar belong to the same partition.

\textbf{Temporal data }are obtained as time-ordered sequences of frames captured during each unloading cycle by fixed high-resolution cameras overlooking the railcar. We segment each sequence into magnet-grab intervals using the magnet position detector and sample one or more keyframes per interval; these per-grab frames constitute the layers in a MIL bag for that railcar.

\textbf{Explainability.} We performed an interpretability analysis of several of our models for a regression task using Grad-CAM ~\cite{Selvaraju_2019}. This is important to verify that our models learn to focus on the scrap-metal region rather than being influenced by production dust or lighting. Figure~\ref{fig:gradcam} shows that transformer architectures produce stronger Grad-CAM activations over the scrap-metal region than convolutional architectures.

\begin{figure*}[t]
\centering
\includegraphics[width=0.9\textwidth]{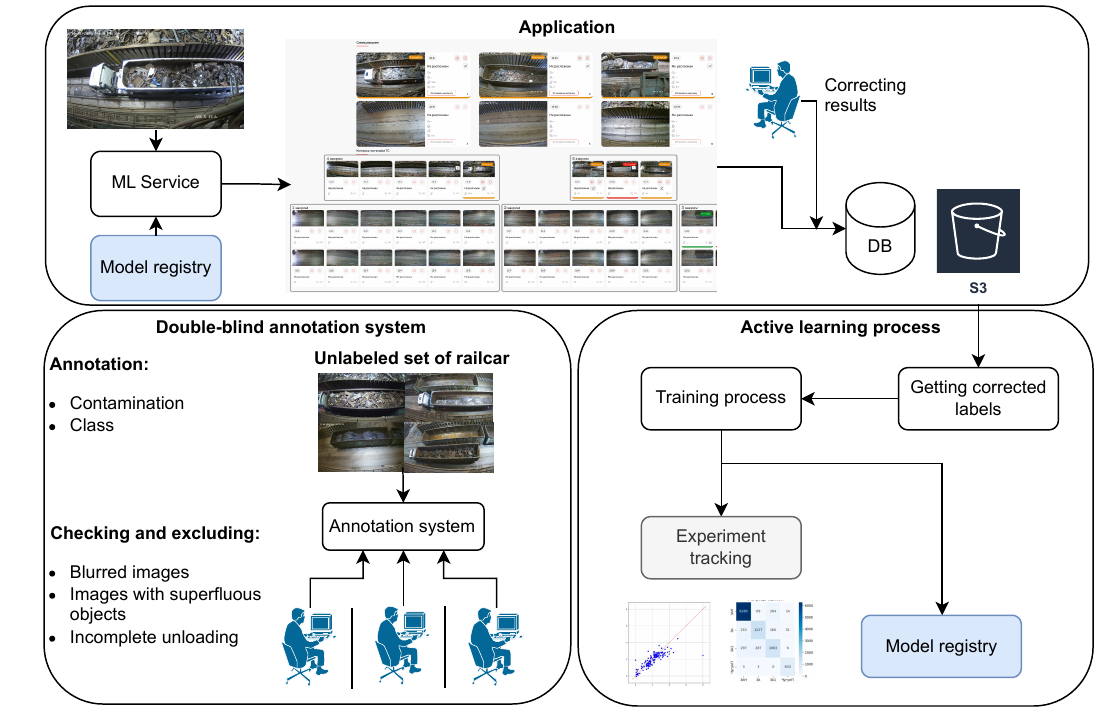} %
\caption{End‑to‑end human‑in‑the‑loop system. \textbf{Double‑blind annotation pipeline} with automated quality checks (blur, extraneous objects, incomplete unloading), triple labeling per railcar, aggregation with dispersion-based adjudication, and audit logging. Production \textbf{application} architecture—magnet/railcar detection and IoU‑based layer (grab) segmentation; MIL pooling with MTL heads for contamination regression and grade classification served via a versioned ML service; results persisted to DB/S3, surfaced to operators for review/correction, and fed back through an \textbf{active‑learning loop} with model registry and experiment tracking.}
\label{human_in_loop}
\end{figure*}
\textbf{MIL} We process each bag of instances with the backbone, compute attention scores over instance embeddings, and take the weighted sum as the bag representation for regression tasks. On top of the aggregated feature, we attach a regression head (Linear($\mathrm{feature}_{dim}$, 256) → ReLU → Dropout → Linear(256, 1)). 
\textbf{Evaluation Strategy.} To determine scrap metal contamination, we use classical regression quality metrics MAE and R2, which are able to adequately estimate the average absolute error and the proportion of the explained variance. This makes it possible to objectively compare models, identify systematic underestimation or overestimation of pollution, and set thresholds for sorting batches.

To determine the scrap metal class, we use Accuracy, Precision, Recall, and F1, which provide a comprehensive assessment of classification quality by balancing precision and recall and minimizing false positives and false negatives-especially important for classes that are similar to each other.

\section{Implementation details}
We keep all backbone architectures intact and only replace their classification layers with identity mappings to expose feature embeddings, upon which we add attention pooling and lightweight task heads. The attention module is a two-layer MLP (Linear–Tanh–Linear) that produces per-instance weights followed by a softmax across instances in a bag.

\textbf{MTL} We share the backbone across tasks and branch into independent task heads that are trained jointly, optionally using the same attention-based bag pooling or a simple mean pooling when bags are present. On top of the aggregated feature, we attach two heads: a regression head (Linear($\mathrm{feature}_{dim}$, 256) → ReLU → Dropout → Linear(256, 1)) and a classification head (Linear($\mathrm{feature}_{dim}$, 256) → ReLU → Dropout → Linear(256, $\mathrm{class}_{num}$)).

\section{Human-in-the-Loop Model Integration}

\subsection{Double-blind annotation system}
The target labels comprise a continuous estimate of non‑metallic contamination and a categorical scrap grade; observations arrive as temporal layers (magnet grabs) represented by one or more quality‑filtered frames. Each railcar is randomly routed to at least three independent annotators under a double‑blind scheme: raters neither see one another’s assessments nor the emerging consensus, and railcar identifiers are pseudonymized to suppress anchoring and social influence. Prior to human review, frames undergo automated eligibility checks that reject corrupted or non‑diagnostic inputs (e.g., checksum failures, abnormal aspect ratios, excessive blur or under/over‑exposure, dust/smoke occlusions, missing railcar occupancy) and filter extraneous objects via a lightweight detector; the UI permits raters to exclude residual low‑quality frames with standardized failure codes (Figure ~\ref{human_in_loop} - Double-blind annotation system). Continuous labels are aggregated by the mean and accompanied by per‑railcar dispersion (standard deviation) as a reliability indicator; items with dispersion above a predefined threshold (e.g., STD 0.4) are sent to adjudication by a senior inspector. Categorical labels are combined by majority vote with expert tiebreak , while per-class confusion is monitored to expose systematic ambiguities. All metadata - rater actions, timestamps, frame eligibility flags, aggregation provenance - are logged with a complete audit trail, and dataset splits are performed at the railcar level to prevent leakage across train, validation and test partitions.

\subsection{Application}

\begin{figure}[!tbp]
\centering
\includegraphics[width=0.8\textwidth]{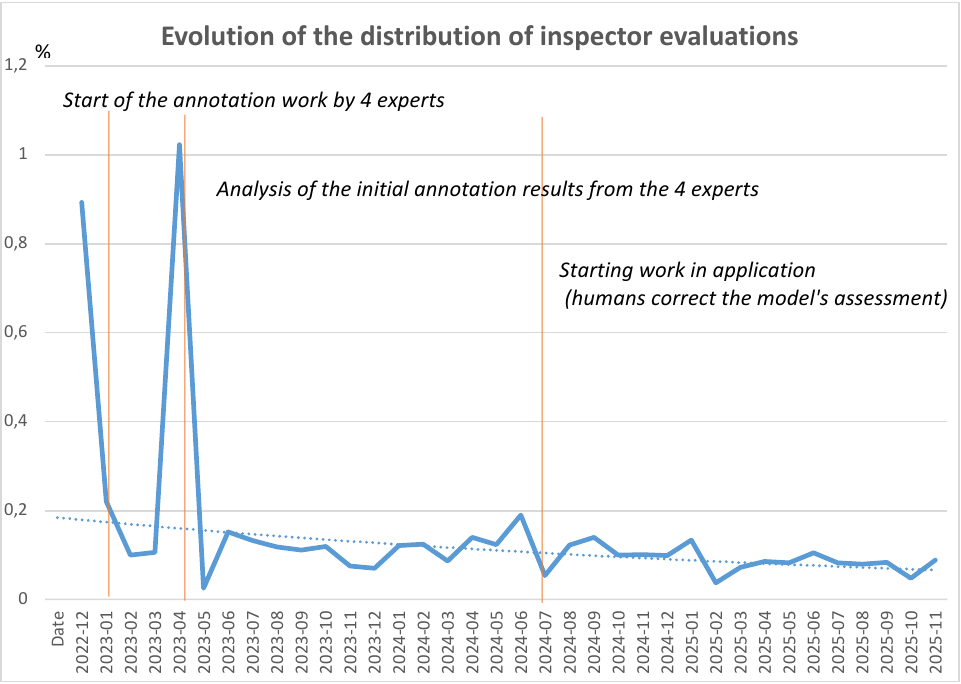}
\caption{With the start of work in the annotation system, inspectors began to more carefully accept scrap metal with minimal discrepancies.}
\label{fig:spread}
\end{figure}

The trained models are deployed within the acceptance application to operate in near real time during railcar unloading (Figure ~\ref{human_in_loop} - Application), with explicit provisions for human oversight and continual improvement. Unloading commences only after spatial conformance is verified by ensuring the railcar centroid lies within a predefined region of interest; a detector–tracker pair localizes the magnet and railcar, and temporal layers are delineated by thresholding the IoU between magnet and railcar masks over time, with keyframes selected near the IoU peak subject to the same quality filters used in annotation. Each layer is submitted to a versioned FastAPI inference service that implements MTL heads for contamination regression and grade classification; outputs include point estimates, confidence scores, and per‑layer features. Results and provenance (railcar ID, timestamps, IoU traces, quality flags, model version) are persisted in a relational store and streamed via a message broker to the operator interface and downstream systems. The plant runs six unloading lines concurrently, with partitioned topics/queues and idempotent writes to guarantee exactly‑once semantics under retries. For every unloading, the system compiles an operational report containing the railcar identifier, contamination percentage, scrap grade, layer counts, confidence summaries, and anomaly/quality flags; latency budgets maintain sub‑second per‑layer inference to keep pace with operations. At the end of unloading, inspectors review and may amend the model’s outputs; edits are captured with rationale codes under role‑based access control, and policy thresholds (e.g., high predicted contamination or low confidence) trigger escalations to manual review. Corrections and uncertain or contentious cases are prioritized for expert re‑labeling and periodically incorporated into versioned datasets. This integration closes the active‑learning loop-linking sensing (Figure ~\ref{human_in_loop} - Active learning process), automated decision support, human validation, and model adaptation-thereby reducing subjective variability, enhancing safety by removing inspectors from hazardous zones, and sustaining calibrated, reliable model performance in production. One of the important features was that with the advent of the annotation system and the work in the application, inspectors began to pay more attention to the acceptance of scrap metal (Figure \ref{fig:spread}).

\section{Discussion}

\textbf{Limitations}
Our approach, while effective in the evaluated setting, has several important limitations. First, the model was trained and validated exclusively on data collected from a single steel recycling facility. This raises concerns about domain generalization: variations in lighting conditions, camera setups, scrap composition, or unloading protocols at other plants may degrade performance.

Second, the ground-truth contamination labels are derived from the average of three human inspector annotations. While this aggregation reduces individual subjectivity and aligns with operational practice, it still inherits the inherent imprecision of visual estimation-inspectors may systematically misjudge contamination due to occlusions, dust, or cognitive bias. 
Consequently, our model learns to approximate a noisy proxy of true contamination rather than an objective physical measurement (e.g., from lab analysis). This limits the theoretical ceiling of achievable accuracy and may propagate human-level errors into automated decisions.

Finally, our current formulation assumes that all layers from a railcar are captured and correctly temporally aligned with magnet movements. In real-world deployments, sensor failures or missed frames could disrupt the MIL bag construction, requiring robust fallback mechanisms.

\textbf{Impact on inspector.} Gradually, inspector tasks-such as photographing scrap metal, filling out reports and being present on the production site-re-convert to the analytical domain: analyze system performance and adjust model outputs to further improve models. At the same time, the demands on reaction speed and other individual human traits are decreasing, thanks to reduced cognitive load and improved working conditions.

\textbf{Future Directions.} Improving the feature space will enable deep learning models to approach or even surpass human-level performance. For instance, incorporating spatial information (e.g., 3D maps to estimate volume) or data from crane scales supporting the electromagnet could provide both density and volume of scrap metal, since non-metallic inclusions are not magnetic.

We plan to publish an industrial dataset that will include data on scrap metal contamination (Figure ~\ref{flow} - Class \& Contamination) and scrap grades from various plants, so that the community can implement mutually beneficial human–machine collaboration in metallurgical production.

\section{Conclusion}

In our research, we proposed the innovative pipline for the define of contamination percentages in scrap metal during the acceptance stage. Our study showed that transformer-based models, specifically the Swin Transformer, significantly outperform traditional CNN architectures in predicting contamination levels from images.

By achieving an MAE of 0.27 and an R2 score of 0.83 at the railcar level (MIL), our proposed method demonstrates high accuracy and reliability, making it suitable for practical deployment in recycling facilities.

The Multi-task approach also shows that it is possible to solve many tasks at once, as inspectors do, and achieve similar quality.

The integration of our pipeline into the industrial process increases the quality of metal scrap delivered to EAF, which subsequently improves the quality of rolled steel products. Inspectors who assess metal scrap are no longer in the danger zone near the unloading area; they now remain safely inside a dedicated room, observing the process remotely.

\section{Acknowledgments}
The work of I. Makarov was supported by the Ministry of Economic Development of the Russian Federation (agreement No. 139-10-2025-034 dd. 19.06.2025, IGK 000000C313925P4D0002)

\bibliographystyle{unsrtnat}

\begin{thebibliography}{10}

\bibitem{linh2019}
Tuan~Linh Dang, Thang Cao, and Yukinobu Hoshino.
\newblock Classification of metal objects using deep neural networks in waste
  processing line.
\newblock {\em IJICIC}, 15(5):1901--1912, October 2019.

\bibitem{balakrishnan2018}
Ramalingam Balakrishnan, Anirudh~Krishna Lakshmanan, Muhammad Ilyas, Anh~Vu Le,
  and Mohan~Rajesh Elara.
\newblock Cascaded machine-learning technique for debris classification in
  floor-cleaning robot application.
\newblock {\em Applied Sciences}, 8(12):2649, 2018.

\bibitem{teguh2021}
Aji~Teguh Prihatno, Ida Bagus Krishna~Yoga Utama, Jun~Yong Kim, and Yeong~Min
  Jang.
\newblock Metal defect classification using deep learning.
\newblock In {\em 2021 Twelfth International Conference on Ubiquitous and
  Future Networks (ICUFN)}, pages 389--393, 2021.

\bibitem{smirnov2021}
Nikolai~V. Smirnov and Aleksey~S. Trifonov.
\newblock Deep learning methods for solving scrap metal classification task.
\newblock In {\em 2021 International Russian Automation Conference
  (RusAutoCon)}, pages 221--225, 2021.

\bibitem{ImpactNMI}
Reinhold Schneider, Valentin Wiesinger, Siegfried Gelder, and Gerhard Reiter.
\newblock Effect of the slag composition on the process behavior, energy
  consumption, and nonmetallic inclusions during electroslag remelting.
\newblock {\em Steel Research International}, 94, April 2023.

\bibitem{smirnov2020}
Nikolai~V. Smirnov and Egor~I. Rybin.
\newblock Machine learning methods for solving scrap metal classification task.
\newblock In {\em 2020 International Russian Automation Conference
  (RusAutoCon)}, pages 1020--1024, 2020.

\bibitem{bircanoglu2018}
Cenk Bircanoğlu, Meltem Atay, Fuat Beşer, Özgün Genç, and Merve~Ayyüce
  Kızrak.
\newblock Recyclenet: Intelligent waste sorting using deep neural networks.
\newblock In {\em 2018 Innovations in Intelligent Systems and Applications
  (INISTA)}, pages 1--7, 2018.

\bibitem{swin2021}
Ze~Liu, Yutong Lin, Yue Cao, Han Hu, Yixuan Wei, Zheng Zhang, Stephen Lin, and
  Baining Guo.
\newblock Swin transformer: Hierarchical vision transformer using shifted
  windows.
\newblock In {\em Proceedings of the IEEE/CVF International Conference on
  Computer Vision}, pages 10012--10022, 2021.

\bibitem{imagenet}
Jia Deng, Wei Dong, Richard Socher, Li-Jia Li, Kai Li, and Li~Fei-Fei.
\newblock Imagenet: A large-scale hierarchical image database.
\newblock In {\em 2009 IEEE Conference on Computer Vision and Pattern
  Recognition}, pages 248--255, 2009.

\bibitem{resnet2015}
Kaiming He, Xiangyu Zhang, Shaoqing Ren, and Jian Sun.
\newblock Deep residual learning for image recognition.
\newblock In {\em 2016 IEEE Conference on Computer Vision and Pattern
  Recognition (CVPR)}, pages 770--778, 2016.

\bibitem{effnet2019}
Mingxing Tan and Quoc Le.
\newblock Efficientnet: Rethinking model scaling for convolutional neural
  networks.
\newblock In {\em International Conference on Machine Learning}, pages
  6105--6114. PMLR, 2019.

\bibitem{vit2020}
Alexey Dosovitskiy, Lucas Beyer, Alexander Kolesnikov, Dirk Weissenborn,
  Xiaohua Zhai, Thomas Unterthiner, Mostafa Dehghani, Matthias Minderer, Georg
  Heigold, Sylvain Gelly, et~al.
\newblock An image is worth 16x16 words: Transformers for image recognition at
  scale.
\newblock In {\em International Conference on Learning Representations}, 2020.

\bibitem{resnext2016}
Saining Xie, Ross Girshick, Piotr Dollar, Zhuowen Tu, and Kaiming He.
\newblock Aggregated residual transformations for deep neural networks.
\newblock In {\em Proceedings of the IEEE Conference on Computer Vision and
  Pattern Recognition (CVPR)}, July 2017.

\bibitem{neu_metal_surface}
Md~Fantacher Islam and Md~Rahman.
\newblock Metal surface defect inspection through deep neural network.
\newblock In {\em International Conference on Computer and Information
  Technology}, 2018.

\bibitem{lin_wu_2024}
Lin Wu.
\newblock Metal surface defect dataset.
\newblock {\em Science Data Bank}, V1, July 2024. \url{https://doi.org/10.57760/sciencedb.10794}

\bibitem{severstal-steel-defect-detection}
Alexey Grishin, BorisV, iBardintsev, inversion, and Oleg.
\newblock Severstal: Steel defect detection.
\newblock Kaggle, 2019.
\newblock \url{https://kaggle.com/competitions/severstal-steel-defect-detection}.

\bibitem{does2023}
Michael Schäfer, Ulrike Faltings, and Björn Glaser.
\newblock Does - a multimodal dataset for supervised and unsupervised analysis
  of steel scrap.
\newblock {\em Scientific Data}, 10:780, November 2023.

\bibitem{10299583}
Zicheng Gao, Hexiao Lu, Jie Lei, Jingbo Zhao, Hao Guo, Chengbing Shi, and Yong
  Zhang.
\newblock An rgb-d-based thickness feature descriptor and its application on
  scrap steel grading.
\newblock {\em IEEE Transactions on Instrumentation and Measurement},
  72:1--14, 2023.

\bibitem{DBLP:journals/corr/abs-1107-2021}
Sivan Sabato and Naftali Tishby.
\newblock Multi-instance learning with any hypothesis class.
\newblock {\em CoRR}, abs/1107.2021, 2011.

\bibitem{10.1023/A:1007379606734}
Rich Caruana.
\newblock Multitask learning.
\newblock {\em Mach. Learn.}, 28(1):41--75, July 1997.

\bibitem{Bozic2021COMIND}
Jakob Bo{\v{z}}i{\v{c}}, Domen Tabernik, and Danijel Sko{\v{c}}aj.
\newblock Mixed supervision for surface-defect detection: from weakly to fully
  supervised learning.
\newblock {\em Computers in Industry}, 2021.

\bibitem{8560423}
Yibin Huang, Congying Qiu, Yue Guo, Xiaonan Wang, and Kui Yuan.
\newblock Surface defect saliency of magnetic tile.
\newblock In {\em 2018 IEEE 14th International Conference on Automation Science
  and Engineering (CASE)}, pages 612--617, 2018.

\bibitem{s20061562}
Xiaoming Lv, Fajie Duan, Jia-jia Jiang, Xiao Fu, and Lin Gan.
\newblock Deep metallic surface defect detection: The new benchmark and
  detection network.
\newblock {\em Sensors}, 20(6):1562, 2020.

\bibitem{DBLP:journals/corr/abs-2103-13003}
Tobias Schlagenhauf, Magnus Landwehr, and J{\"{u}}rgen Fleischer.
\newblock Industrial machine tool component surface defect dataset.
\newblock {\em CoRR}, abs/2103.13003, 2021.

\bibitem{doi:10.1080/08839514.2021.1975391}
Dingming Yang, Yanrong Cui, Zeyu Yu, and Hongqiang Yuan.
\newblock Deep learning based steel pipe weld defect detection.
\newblock {\em Applied Artificial Intelligence}, pages 1--13, 2021.

\bibitem{nature_defects}
Yang Gao, Gang Lv, Dong Xiao, Xize Han, Tao Sun, and Zhenni Li.
\newblock Research on steel surface defect classification method based on deep
  learning.
\newblock {\em Scientific Reports}, 14, April 2024.

\bibitem{auer2019artificialintelligencebasedprocessmetal}
Maximilian Auer, Kai Osswald, Raphael Volz, and Joerg Woidasky.
\newblock Artificial intelligence-based process for metal scrap sorting.
\newblock {\em arXiv preprint arXiv:1903.09415}, 2019.

\bibitem{Selvaraju_2019}
Ramprasaath~R. Selvaraju, Michael Cogswell, Abhishek Das, Ramakrishna Vedantam,
  Devi Parikh, and Dhruv Batra.
\newblock Grad-cam: Visual explanations from deep networks via gradient-based
  localization.
\newblock {\em International Journal of Computer Vision}, 128(2):336--359,
  October 2019.

\bibitem{nmis2021}
Andrey Zhitenev, Maria Salynova, Alexey Shamshurin, Sergey Ryaboshuk, and
  Vladislav Kolnyshenko.
\newblock Database clustering after automatic feature analysis of nonmetallic
  inclusions in steel.
\newblock {\em Metals}, 11(10):1650, 2021.

\bibitem{klimek2024circulartransformationeuropeansteel}
Peter Klimek, Maximilian Hess, Markus Gerschberger, and Stefan Thurner.
\newblock Circular transformation of the European steel industry renders scrap
  metal a strategic resource.
\newblock {\em arXiv preprint arXiv:2406.12098}, 2024.

\bibitem{cis2018}
A.~A. Kazakov and A.~I. Zhitenev.
\newblock Assessment and interpretation of nonmetallic inclusions in steel.
\newblock {\em CIS Iron and Steel Review}, pages 33--38, October 2018.

\end{thebibliography}

\end{document}